%% file: EPIC.tex
\documentclass[10pt,twocolumn,letterpaper]{article}

\usepackage{cvpr}              

\usepackage{graphicx}
\usepackage{amsmath}
\usepackage{amssymb}
\usepackage{booktabs}
\usepackage{color}
\usepackage{colortbl}
\usepackage[dvipsnames]{xcolor}
\usepackage{float}
\usepackage{multirow}
\usepackage{algorithm}
\usepackage[frozencache,cachedir=.]{minted}
\usepackage[pagebackref,breaklinks,colorlinks]{hyperref}
\usepackage{graphicx}
\usepackage{pythonhighlight}

\usepackage[capitalize]{cleveref}
\crefname{section}{Sec.}{Secs.}
\Crefname{section}{Section}{Sections}
\Crefname{table}{Table}{Tables}
\crefname{table}{Tab.}{Tabs.}

\def\cvprPaperID{*****} 
\def\confName{CVPR}
\def\confYear{2022}
\def\cvprPaperID{*****} 
\def\confName{CVPR}
\def\confYear{2022}

\begin{document}

\newcommand{\vlp}{Egocentric VLP}
\newcommand{\dataset}{EgoClip}
\newcommand{\model}{EgoNCE\xspace}
\newcommand{\eval}{EgoMCQ}
\newcommand{\epic}{EPIC-KITCHENS-100}
\newcommand{\charades}{Charades-Ego}
\newcommand{\lta}{video-text localization}
\newcommand{\mir}{Multi-Instance Retrieval}
\newcommand{\nlq}{Natural Language Query}
\newcommand{\mq}{Moment Query}
\newcommand{\ossc}{Object State Change Classification}
\newcommand{\web}{WebVid-2M}
\newcommand{\ccweb}{CC3M+WebVid-2M}
\newcommand{\howto}{HowTo100M}

\title{Egocentric Video-Language Pretraining~ \\ @~\epic~\mir~Challenge~2022}  

\author{Kevin Qinghong Lin$^1$\quad Alex Jinpeng Wang$^1$\quad Rui Yan$^1$\quad Eric Zhongcong Xu$^1$\quad Rongcheng Tu$^2$\\
Yanru Zhu$^2$\quad Wenzhe Zhao$^2$\quad Weijie Kong$^2$\quad Chengfei Cai$^2$\quad Hongfa Wang$^2$\\
Wei Liu$^2$\quad Mike Zheng Shou$^1$\thanks{Corresponding Author.}\\
\\
$^1$Show Lab, National University of Singapore\quad$^2$Tencent Data Platform\\
{\tt\small \{kevin.qh.lin, yanrui6019, turongcheng\}@gmail.com, 
\{jinpengwang, zhongcongxu\}@u.nus.edu}\\
\tt\small \{yizhizhu, carsonzhao, jacobkong, fletchercai, hongfawang\}@tencent.com\\
\tt\small wl2223@columbia.edu, mike.zheng.shou@gmail.com
}
\maketitle

\begin{abstract}
In this report, we propose a video-language pretraining~(VLP) based solution~\cite{kevin2022egovlp} for the \epic~\mir~(MIR) challenge. Especially, we exploit the recently released Ego4D dataset~\cite{grauman2021ego4d} to pioneer \vlp~from pretraining dataset, pretraining objective, and development set. Based on the above three designs, we develop a pretrained video-language model that is able to transfer its egocentric video-text representation to MIR benchmark. Furthermore, we devise an adaptive multi-instance max-margin loss to effectively fine-tune the model and equip the dual-softmax technique for reliable inference. Our best single model obtains strong performance on the challenge test set with $47.39\%$ mAP and $61.44\%$ \% nDCG. The code is available at {\url{https://github.com/showlab/EgoVLP}}.
\end{abstract}

\section{Introduction}\label{sec:intro}
Video-Language Pretraining~(VLP) has prevailed in the regime of Vision~+~Language, aiming to learn strong and transferable video-language representation for powering a broad spectrum of video-text downstream tasks, video-text retrieval, video question answering, video-captioning.
The successes of VLP mainly stems from the availability of large-scale open-world video-text datasets such as \howto~\cite{miech2019howto100m}, which scrapes $134$K hours of instructional videos from the YouTube accompanied by text yielded from Automatic Speech Recognition.

Despite reaching an impressive data scale, videos in the existing video-text pretraining datasets~\cite{miech2019howto100m, bain2021frozen} are often of 3rd-person views and might have been edited before posting on the web.
Yet, there is a noticeable domain gap between the existing video-text pretraining datasets and 1st-person view videos such as those videos captured by wearable cameras or smart glasses.
Egocentric video has received increasing interests from academia~(e.g., activity anticipation~\cite{damen2022rescaling}) and industry (various applications in robotics and augmented reality).
But, due to such a domain gap, directly transferring the existing VLP models to egocentric downstream tasks cannot fully unleash the potential of large-scale pretraining approaches.
Roused by the favorable scale and diversity of recently released Ego4D~\cite{grauman2021ego4d} dataset, we are motivated to develop \vlp~models~\cite{kevin2022egovlp}, which can greatly benefit various egocentric video downstream applications.

In this report, we leverage our \vlp~\cite{kevin2022egovlp}~for powering \epic~\mir~(MIR)~challenge. 
We provide a comprehensive analysis of the impact of different VLPs on this task, e.g., without VLP, 3rd-person VLP, and 1st-person VLP.
Furthermore, to effectively transfer the video-text representation to MIR~task, we devise an adaptive multi-instance maxmargin loss for fine-tuning.
Besides, we introduce the dual-softmax technique for reliable inference.

\section{Approach}\label{sec:method}
\begin{figure*}[!t]
	\centering
	\includegraphics[width=1.0\linewidth]{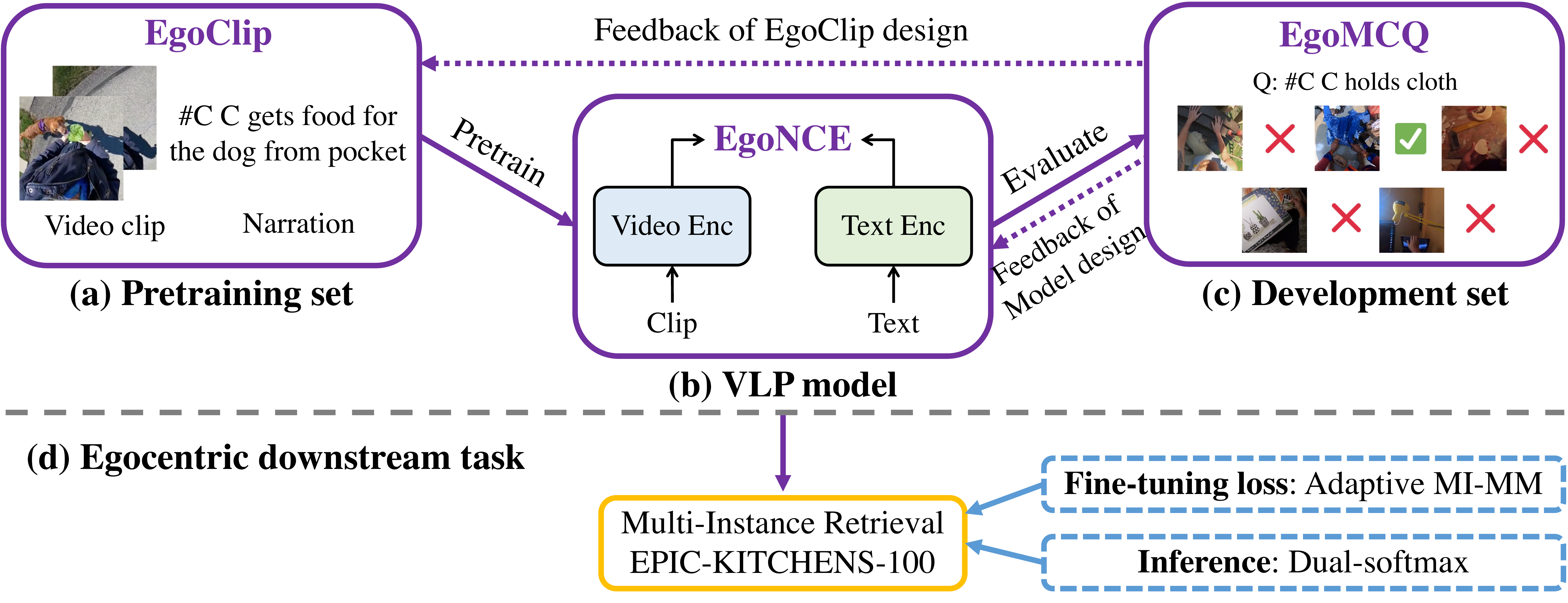}
	\vspace{-.5em}
	\caption{
	\textbf{Top:} Our \vlp~framework, which includes: (a) pretraining set \dataset; (b) VLP model; and (c) development set \eval. We use \dataset~to pretrain a VLP model with \model~ loss and then evaluate on \eval. According to the feedback, we iteratively refine our designs of (a) and (b). 
	\textbf{Down:}
	We transfer our pretrained model to \epic~\mir~task by equipping two techniques: the adaptive multi-instance max-margin loss for fine-tuning and the dual-softmax for inference.}
	\vspace{-1.5em}
	\label{framework}
\end{figure*}

\subsection{VLP Model}
We choose Frozen~\cite{bain2021frozen} as our pretraining architecture. As depicted in the Fig.~\ref{framework}(b), Frozen~\cite{bain2021frozen} design encompasses an elegant and simple dual encoder strategy (one per modality) which has favorable characteristics (e.g., indexability and efficiency~\cite{bain2021frozen}). 
Note that this allows the pretrained model for single-modality tasks (e.g., video-only tasks). 
In practice, the video encoder is a TimeSformer~\cite{timesformer} architecture while the text encoder builds upon DistillBERT~\cite{distilbert}. 
We adopt this notation: $(\mathcal{V}_i, \mathcal{T}_i)$ represent the video-test input to the model, while $\mathbf{v}_i$ and $\mathbf{t}_i$ are used to identify the L2 normalized video and text embedding with $d$ dimension.

\subsection{Egocentric Pretraining}
As illustrated in Fig.~\ref{framework}~\cite{kevin2022egovlp}, our pretraining framework includes three designs: \dataset, \model, and \eval.
We use \dataset~dataset for pretraining, which comprises $3.85$M video-text pairs well-chosen from Ego4D, covering a large variety of human daily activities. Details please refer to~\cite{kevin2022egovlp}.
Next, we employ \model~as the model pretraining objective, which extends video-text InfoNCE~\cite{bain2021frozen} via positive and negative sampling strategies with formulation:
\begin{equation}
\mathcal{L}^\text{ego}=
\mathcal{L}^\text{ego}_\text{v2t}
+\mathcal{L}^\text{ego}_\text{t2v}.
\end{equation}

We formulate $\mathcal{L}^\text{ego}_\text{v2t}$ for simplicity whereas $\mathcal{L}^\text{ego}_\text{t2v}$ is defined in a symmetry way.
\begin{equation}
\begin{split}
	\mathcal{L}^\text{ego}_\text{v2t}=\frac{1}{| \mathcal{\widetilde{B}} |}\sum_{i\in\mathcal{\widetilde{B}}}  \log 
	\frac{
	{
	\sum_{k\in \mathcal{P}_i}\exp(\mathbf{v}_i^T\mathbf{t}_k /\tau)
	}
	}
	{  \sum_{j\in \mathcal{B}} \left( \exp(\mathbf{v}_i^T\mathbf{t}_j /\tau) +
	{\exp(\mathbf{v}_i^T\mathbf{t}_{j'} /\tau)}  \right) 
	},
	\label{egonce}
\end{split}
\end{equation}
where the numerator term corresponds to our proposed \textbf{action-aware positive samples}, which select the positive sample within a batch by identifying narrations nouns and verbs. Then, batch samples that shared at least one noun and at least one verb are treated as positive samples:
$\mathcal{P}_i=\{j\in \mathcal{B}~|~\text{noun}(j)\cap\text{noun}(i)\neq\varnothing, \text{verb}(j)\cap\text{verb}(i)\neq\varnothing\}$.
While the denominator term corresponds to our proposed \textbf{scene-aware negative samples}.
For each video clip $i$, we sample an adjacent clip $i'\in \mathcal{N}(i)$, which is close to $i$ in time~(less than 1 min) within the same video.
Hence the batch is updated as $\mathcal{\widetilde{B}}=\{\underbrace{1,2,\cdots N}_{\mathcal{B}}, \underbrace{1',2',\cdots, N'}_{\mathcal{N}(\mathcal{B})} \}$. 
\model provides a general extension to adapt the existing VLP models for video-text pretraining datasets in the egocentric domain.

We evaluate our designs of \dataset~and \model on \eval, 
which contains 39K video-text multi-choices questions that are closer to pretraining domains and benchmark model video-text alignment, powering us to accurately validate and quickly iterate our decisions.

\input{tex/epic_sota}

\subsection{Task-specific Transferring}
In this section, we focus on effectively transferring pretrained video-text representations to \epic~\mir~task. 
In this task, a narration may be jointly associated with multiple clips, so a multi-instance learning mechanism can better handle such a situation. And this dataset provides the action label to calculate the correlation $c_{ij}\in [0,1]$ between two clip-text pairs $(i,j)$, which supports the application of Multi-Instance MaxMargin loss~(MI-MM), as recommended in baseline~\cite{wray2019fine}. 
\begin{equation}
\begin{aligned}
\mathcal{L}=
\sum_{(i,j,k)\in \Omega}\max
\left( \gamma + \mathbf{v}_i^T\mathbf{t}_j
-\mathbf{v}_i^T \mathbf{t}_k \right)
+
\left( \gamma + \mathbf{t}_i^T\mathbf{v}_j
-\mathbf{t}_i^T \mathbf{v}_k \right),
\label{mimm}
\end{aligned}
\end{equation}
where $\Omega=\{(i,j,k)~|j\in i^{+}, k\in i^{-}\}$ is a triple, which indicates a positive instance $j$ and a negative instance $k$ for $i$. 
In our setting, we define the positive set as $i^{+}=\{j|c_{ij}>0.1\}$ and the negative as the remains sample within the batch. The $\gamma$ is a constant margin factor.

However, different combinations are shared with the same margin $\gamma$ in Eq.~\ref{mimm} and thus are treated equally when fine-tuning.
Intuitively, if two sample $(i,j)$ are highly similar, they should be pulled closer with a larger margin surpassing the $(i,k)$. Otherwise, they should be pulled with a small margin if not very similar. Thus, we devise the following \textbf{Adaptive MI-MM} to extend the Eq.\ref{mimm}.
\begin{equation}
\begin{aligned}
\mathcal{L}^{\dagger}=
\sum_{(i,j,k)\in \Omega}\max
\left( c_{ij}\gamma + \mathbf{v}_i^T\mathbf{t}_j
-\mathbf{v}_i^T \mathbf{t}_k \right)
+\\
\left( c_{ij}\gamma + \mathbf{t}_i^T\mathbf{v}_j
-\mathbf{t}_i^T \mathbf{v}_k \right),
\label{ada_mimm}
\end{aligned}
\end{equation}
where $c_{ij}$ adaptively control the marginal, e.g.,  two instances $(i,j)$ that are semantically identical~($c_{ij}=1$)  will be assigned a largest marin $1.0\gamma$. Otherwise, a less margin $0.1\gamma$ is given when they are not very similar~($c_{ij}=0.1$).

\textbf{Inference.}
After we finalize the fine-tuning, we use the model to encode video and text embeddings for all samples within the test set.
To obtain the cross-modal retrieval results, a common way is to calculate the similarity score between a text embedding $\mathbf{t}_i$ and a video embedding $\mathbf{v}_j$ and index the maximum as the top retrieval result.
Here, motivated by~\cite{cheng2021improving}, we introduce the dual softmax techniques to better scale the similarities and filter the hard case, thus reaching more reliable prediction results.
We show the PyTorch-like pseudo-code in Alg.~\ref{dual_softmax} to compare the two inference way. Notably, the dual-softmax only works on inference and thus does not introduce additional training costs, and it is flexible to different models.

\begin{algorithm}[h]
\caption{Pseudo-code for Dual-softmax (PyTorch-like)}
\begin{python}
# Input(embeddings): T_{Nxd}, V_{Mxd}
# Output(scores): res_{NxM}

# (1) the common way
sim = torch.mm(T,V)
res = F.softmax(sim, axis=0)

# (2) dual-softmax
sim = torch.mm(T,V)
prior = F.softmax(sim/500, axis=1)
res = F.softmax(prior*sim, axis=0)
\end{python}
\label{dual_softmax}
\vspace{-1.em}
\end{algorithm}

\section{Experiments}\label{sec:exps}
\subsection{Implementation Details}
Following the settings of official Frozen~\cite{bain2021frozen}\footnote{https://github.com/m-bain/frozen-in-time}, the video encoder is initialized with ViT~\cite{dosovitskiy2020image} weights trained on ImageNet-21K with sequence dimension $D=768$. 
The text encoder is based on huggingface's $\texttt{distilbert-base-uncased}$.
The dimension of common feature space is set as $256$, and the temperature parameter $\tau$ is set to $0.05$.
During pretraining, each video is resized to $224\times 224$ as input with sample frames number $4$ and batch size $512$.
We use the Adam optimizer with a learning rate of $3\times 10^{-5}$ with a total epoch of $10$. 
When transferring to MIR~task, we select the checkpoints with the best score on \eval~benchmark and fine tune the VLP model on the MIR~training set with $67.2$K clips.
We set the training epoch as $100$ and keep other settings the same as pretraining.
In the next Sec.~\ref{exp_pretraining}, we use the MI-MM loss with $\gamma$ equal to $0.2$ for fine-tuning. And we validate our proposed Adaptive MI-MM and dual-softmax in Sec.~\ref{exp_ablation}. 
Since most correlation $c_{ij}$ equal $0.5$ in the MIR dataset, we double the $\gamma$ of Adaptive MI-MM to $0.4$ to align with the margin of vanilla MI-MM loss.

\subsection{Pretraining Effects}\label{exp_pretraining}
In Tab.~\ref{mir}, we report both zero-shot and fine-tuning evaluation results of different VLP.
In the zero-shot setting, 
pretraining with \dataset~($3.8$M),~despite being smaller in scale, still outperforms  \ccweb~($5.5$M)~and \howto~($136$M), validating the unique benefit of pretraining on egocentric data.
When fine-tuned with $4$ frames, \dataset~pretraining  maintains a margin over the best baseline \ccweb, further verifying the viewpoint domain gap within fine-tuning.
Lastly, we increase the sample frames of our finalized model as well as best competitor \ccweb~pretraining to $16$.
As expected, performance gains accompany the frame increase. 
We deem that notable benefits come from better temporal modeling for frequent action interactions in the 1st-person view. 
Overall, our pretraining model outperforms the best baseline (JPoSE) by $1.0$ mAP and $5.9\%$ nDCG while requiring fewer frames and input modalities.\\
\input{tex/epic_curve}

In Fig~\ref{fig_epic_curve}, we display training curves of MIR under different VLP discussed in Tab.\ref{exp_pretraining}.
We can found that:
These models with video-text pretraining have a faster rise in performance. Except for \howto, which is close to baseline without pretraining. With \dataset~for egocentric pretraining, the VLP model achieves nearly convergent performance with only a small number of epochs~(less than $20$). Especially with \model~as the pretraining objective, this positive effect is further enhanced.

\subsection{Transferring Ablations}\label{exp_ablation}
In Tab.\ref{mir_abla}, we validate different fine-tuning strategies when transfer the best pretrained model~(Frozen+EgoNCE in Tab.\ref{mir}) to \mir~task, and we adopt the common way to calculate the similarity scores by default.
It shows that InfoNCE performs poorly as a fine-tuning loss despite it being widely used in 3rd-person datasets e.g., Frozen~\cite{bain2021frozen} fine-tune on MSR-VTT.
When replacing InfoNCE with MI-MM~(Eq.\ref{mimm}), there is a signifcant improvement, since MI-MM is well aligned with the mutli-positive characteristic of the \epic.
Moreover, Adaptive MI-MM pushes the performance beyond MI-MM by introducing an adaptive margin~(Eq.\ref{ada_mimm}), thus serving as a better fine-tune objective in MIR.
By equipping dual-softmax to scale similarities, we reach extra $1.2\%$ mAP and $1.0$ nDCG performance gains, which is our best single-model performance.

\input{tex/epic_abla}

\section{Conclusion and Limitations}
We present an egocentric video-language pretraining solution~\cite{kevin2022egovlp} for the \epic~MIR~challenge. 
Specifically, we develop a video-language transformer model and exploit the recently released Ego4D dataset~\cite{grauman2021ego4d} to reach strong video-text representation.
Furthermore, for this challenge, we devise an Adaptive MI-MM loss to fine-tune and adopt dual-softmax techniques to improve inference.
Extensive experimental results validate the effectiveness of our \vlp~and the transferring strategies.

\textbf{Limitations:}
VLP requires a large training cost ($1,536$ GPU hrs for our model) and may be limited by the model architecture thus not flexible for a specific task.

{\small
\bibliographystyle{ieee_fullname}
\bibliography{EPIC}
}

\end{document}

%% file: tex/epic_sota.tex
\begin{table*}[t]
\small
\centering
\resizebox{0.95\textwidth}{!}
{%
\begin{tabular}{lccc|cccccc}
    \toprule[1pt]
    \multirow{2}{*}{\textbf{Methods}} & \multirow{2}{*}{\textbf{Vis Enc Input}} &\multirow{2}{*}{\textbf{\# Frames}}  & \multirow{2}{*}{\textbf{Vis-text PT}} & \multicolumn{3}{c}{\textbf{mAP~(\%)}} & \multicolumn{3}{c}{\textbf{nDCG~(\%)}} \\
    &  & & & V$\rightarrow$T & T$\rightarrow$V & Avg & V$\rightarrow$T & T$\rightarrow$V  & Avg \\ 
    \midrule[1pt] 
    Random & - &-&-                                                             & $5.7$  & $5.6$  & $5.7$  & $10.8$ & $10.9$ & $10.9$. \\ 
    MI-MM& S3D & $32$ & HowTo100M                      & $34.8$ & $23.6$ & $29.2$ & $47.1$ & $42.4$ & $44.7$  \\
    MME~\cite{wray2019fine}   & TBN~$\dagger$~\cite{kazakos2019epic} & $25$ & - & $43.0$ & $34.0$ & $38.5$ & $50.1$ & $46.9$ & $48.5$  \\
    JPoSE~\cite{wray2019fine} & TBN~$\dagger$~\cite{kazakos2019epic} & $25$ & - & $49.9$ & $38.1$ & $44.0$ & $55.5$ & $51.6$ & $53.5$  \\ 
    \midrule
    Frozen~ & Raw Videos& $4$ & -             & $38.8$ & $29.7$ & $34.2$ & $50.5$ & $48.3$ & $49.4$ \\
    Frozen~ & Raw Videos& $4$ & HowTo100M     & $39.2$ & $30.1$ & $34.7$ & $50.7$ & $48.7$ & $49.7$ \\
    Frozen~ & Raw Videos& $4$ & CC3M+WebVid2M & $41.2$ & $31.6$ & $36.4$ & $52.7$ & $50.2$ & $51.4$ \\
    Frozen~ & Raw Videos& $4$ &  \dataset     & \underline{$44.5$} & \underline{$34.7$} & \underline{$39.6$} & \underline{$55.7$} & \underline{$52.9$} & \underline{$54.3$} \\ 
    Frozen+\model~& Raw Videos& $4$ & \dataset & $\mathbf{45.1}$ &  $\mathbf{35.3}$ & $\mathbf{40.2}$ & $\mathbf{56.2}$ & $\mathbf{53.5}$ & $\mathbf{54.8}$\\ 
    \midrule[1pt]
    Frozen~ & Raw Videos& 16 & CC3M+WebVid2M&  \underline{$45.8$} & \underline{$36.0$}  & \underline{$40.9$}  & \underline{$57.2$} & \underline{$54.3$} & \underline{$55.8$}  \\
    Frozen+\model~& Raw Videos& 16& \dataset & $\mathbf{49.9}$ &  $\mathbf{40.5}$ &$\mathbf{45.0}$ & $\mathbf{60.9}$ & $\mathbf{57.9}$ & $\mathbf{59.4}$\\
    \midrule[1pt]
    \rowcolor[gray]{0.9}
    Frozen~ & Raw Videos& $4$ & HowTo100M.      & $6.8$ & $6.3$ & $6.5$ & $11.6$ & $12.8$ & $12.2$  \\
    \rowcolor[gray]{0.9}
    Frozen~ & Raw Videos& $4$ & CC3M+WebVid2M.  & $8.6$ & $7.4$ & $8.0$ & $14.5$ & $14.6$ & $14.5$  \\
    \rowcolor[gray]{0.9}
    Frozen~& Raw Videos& 4& \dataset.           & \underline{$17.9$} & \underline{$13.1$} & \underline{$15.5$} & \underline{$23.0$} & \underline{$21.2$} & \underline{$22.1$}  \\ 
    \rowcolor[gray]{0.9}
    Frozen+\model~& Raw Videos& $4$& \dataset& $\mathbf{19.4}$ & $\mathbf{13.9}$ & $\mathbf{16.6}$ &  $\mathbf{24.1}$ & $\mathbf{22.0}$ &$\mathbf{23.1}$  \\
    \bottomrule[1pt] 
\end{tabular}
}
\centering
\vspace{0em}
\caption{
Performance of the \epic~Multi-Instance Retrieval.
Note that TBN~$\dagger$ feature~\cite{kazakos2019epic} are a combination of three modalities: RGB, Flow and Audio.
Conversely, our approach only relies on RGB input. 
The \colorbox{gray!20}{\makebox(30,4){grey rows}} correspond to \textbf{zero-shot evaluation}.
}
\vspace{-1.5em}
\label{mir}
\end{table*}

%% file: tex/epic_curve.tex
\begin{figure}[h] \centering
    \begin{subfigure}{0.49\linewidth}
        \includegraphics[width=1.0\linewidth]{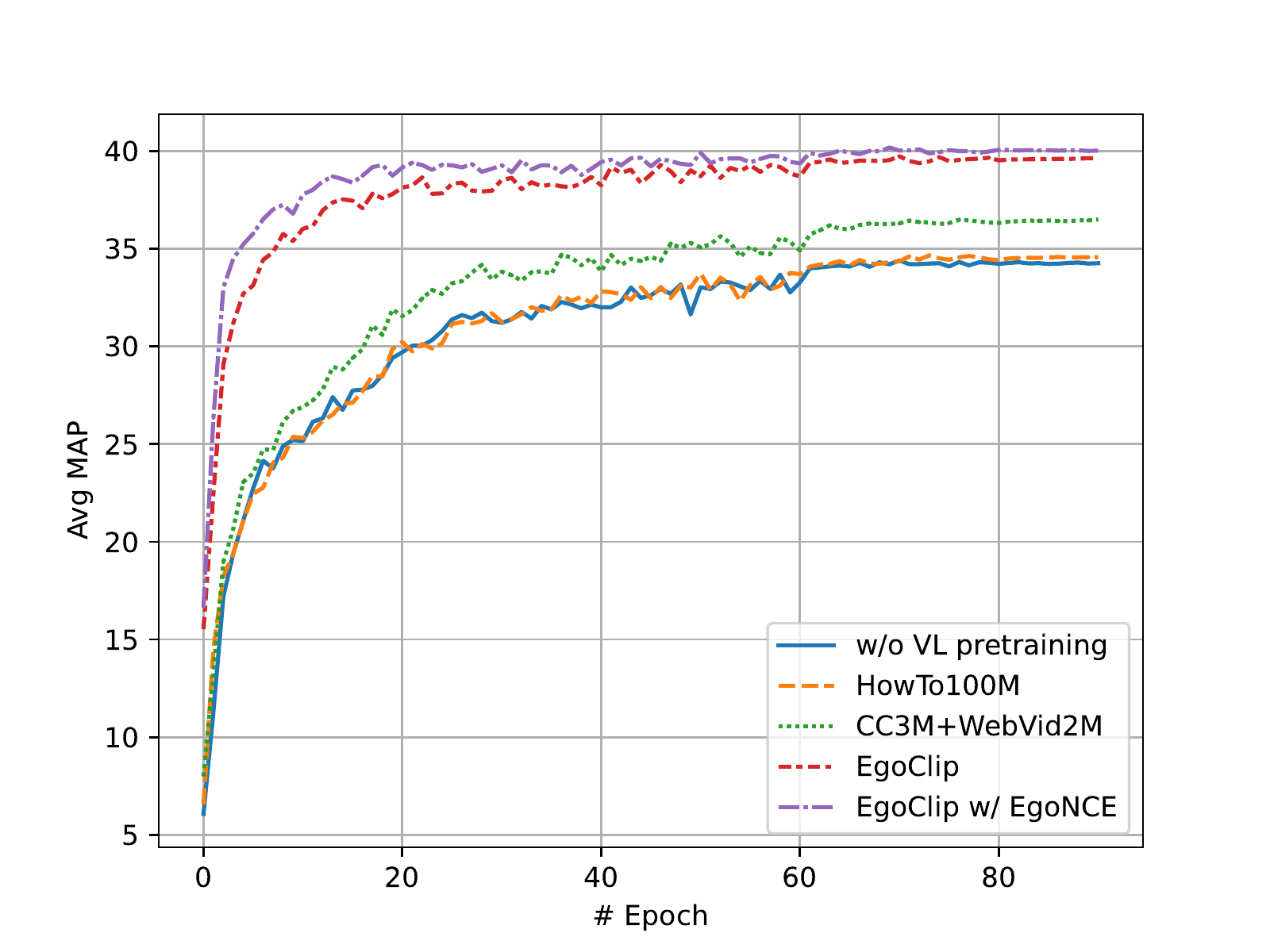}
        \caption{mAP with training epoch}
    \end{subfigure} %
    \begin{subfigure}{0.49\linewidth}    
        \includegraphics[width=1.0\linewidth]{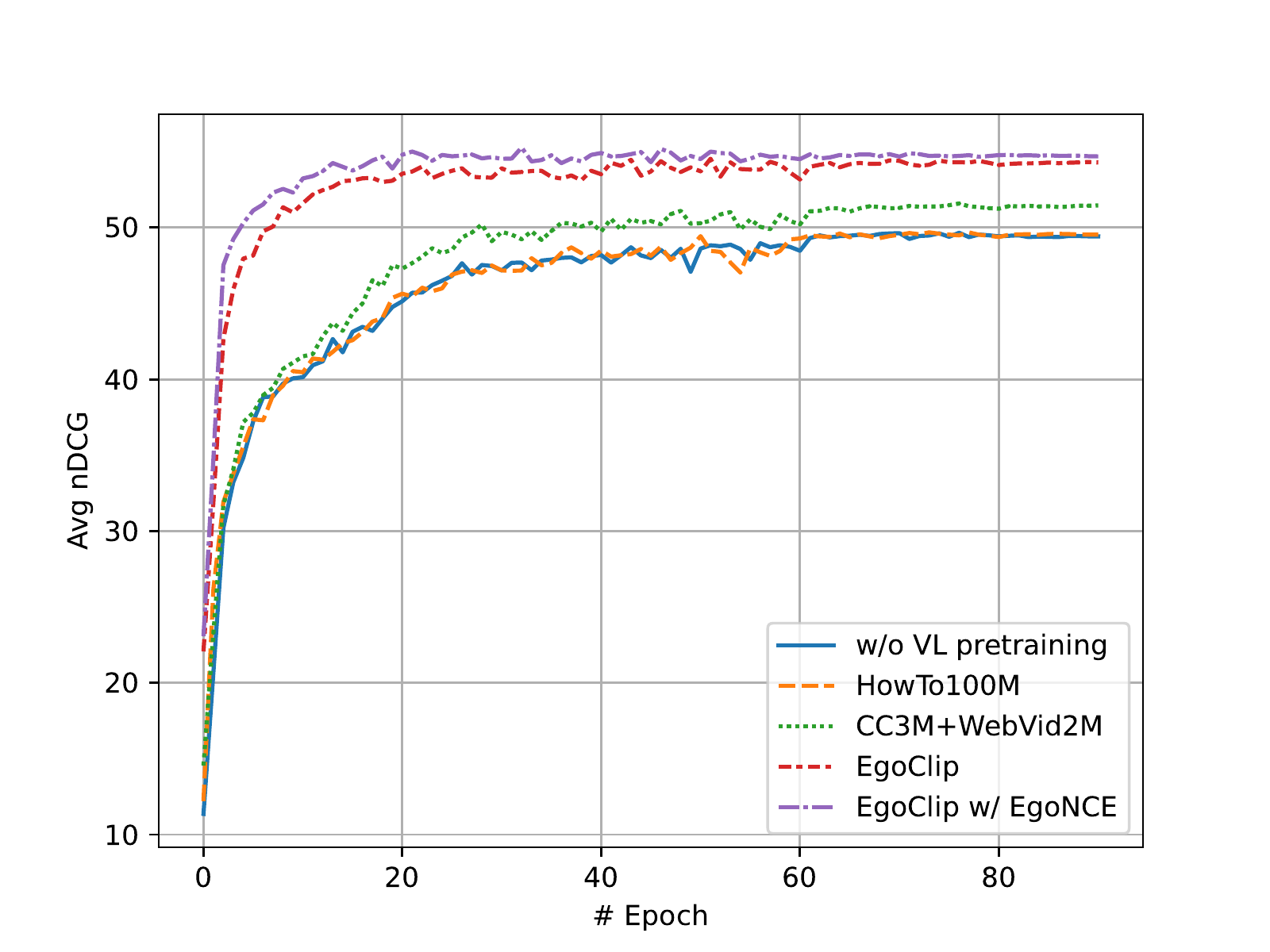}
        \caption{nDCG with training epoch}
    \end{subfigure} 
    \caption{Training curves of MIR task.}
    \label{fig_epic_curve}
\end{figure}

%% file: tex/epic_abla.tex
\begin{table}[t]
\small
\centering
\resizebox{0.48\textwidth}{!}
{%
\begin{tabular}{l|cccccc}
    \toprule[1pt]
    \multirow{2}{*}{\textbf{Methods}} & \multicolumn{3}{c}{\textbf{mAP~(\%)}} & \multicolumn{3}{c}{\textbf{nDCG~(\%)}} \\
    & V$\rightarrow$T & T$\rightarrow$V & Avg & V$\rightarrow$T & T$\rightarrow$V  & Avg \\ 
    \midrule[1pt] 
    InfoNCE & $40.9$ & $34.9$ & $37.9$ & $57.8$ & $56.0$ & $56.9$   \\
    MI-MM & $49.9$ & $\mathbf{40.5}$ & $45.0$ & $60.9$ & $57.9$ & $59.4$ \\
    Adaptive MI-MM & $\mathbf{52.3}$ & ${40.1}$ & $\mathbf{46.2}$ & $\mathbf{62.2}$ & $\mathbf{58.6}$ & $\mathbf{60.4}$\\
    \midrule[1pt] 
    w/ Dual softmax & $\mathbf{53.8}$ & $\mathbf{40.9}$ & $\mathbf{47.4}$ & $\mathbf{63.2}$ & $\mathbf{59.6}$ & $\mathbf{61.4}$\\
    \bottomrule[1pt]
\end{tabular}
}
\centering
\caption{Ablation of different transferring strategies.}
\vspace{-1em}
\label{mir_abla}
\end{table}

%% file: EPIC.bbl
\begin{thebibliography}{10}\itemsep=-1pt

\bibitem{bain2021frozen}
Max Bain, Arsha Nagrani, G{\"u}l Varol, and Andrew Zisserman.
\newblock Frozen in time: A joint video and image encoder for end-to-end
  retrieval.
\newblock In {\em ICCV}, pages 1728--1738, 2021.

\bibitem{timesformer}
Gedas Bertasius, Heng Wang, and Lorenzo Torresani.
\newblock Is space-time attention all you need for video understanding.
\newblock {\em arXiv preprint arXiv:2102.05095}, 2(3):4, 2021.

\bibitem{cheng2021improving}
Xing Cheng, Hezheng Lin, Xiangyu Wu, Fan Yang, and Dong Shen.
\newblock Improving video-text retrieval by multi-stream corpus alignment and
  dual softmax loss.
\newblock {\em arXiv preprint arXiv:2109.04290}, 2021.

\bibitem{damen2022rescaling}
Dima Damen, Hazel Doughty, Giovanni~Maria Farinella, Antonino Furnari,
  Evangelos Kazakos, Jian Ma, Davide Moltisanti, Jonathan Munro, Toby Perrett,
  Will Price, et~al.
\newblock Rescaling egocentric vision: Collection, pipeline and challenges for
  epic-kitchens-100.
\newblock {\em IJCV}, 130(1):33--55, 2022.

\bibitem{dosovitskiy2020image}
Alexey Dosovitskiy, Lucas Beyer, Alexander Kolesnikov, Dirk Weissenborn,
  Xiaohua Zhai, Thomas Unterthiner, Mostafa Dehghani, Matthias Minderer, Georg
  Heigold, Sylvain Gelly, et~al.
\newblock An image is worth 16x16 words: Transformers for image recognition at
  scale.
\newblock In {\em International Conference on Learning Representations}, 2020.

\bibitem{grauman2021ego4d}
Kristen Grauman, Andrew Westbury, Eugene Byrne, Zachary Chavis, Antonino
  Furnari, Rohit Girdhar, Jackson Hamburger, Hao Jiang, Miao Liu, Xingyu Liu,
  et~al.
\newblock Ego4d: Around the world in 3,000 hours of egocentric video.
\newblock {\em arXiv preprint arXiv:2110.07058}, 2021.

\bibitem{kazakos2019epic}
Evangelos Kazakos, Arsha Nagrani, Andrew Zisserman, and Dima Damen.
\newblock Epic-fusion: Audio-visual temporal binding for egocentric action
  recognition.
\newblock In {\em ICCV}, pages 5492--5501, 2019.

\bibitem{kevin2022egovlp}
Kevin~Qinghong Lin, Alex~Jinpeng Wang, Mattia Soldan, Michael Wray, Rui Yan,
  Eric~Zhongcong Xu, Difei Gao, Rongcheng Tu, Wenzhe Zhao, Weijie Kong, et~al.
\newblock Egocentric video-language pretraining.
\newblock {\em arXiv preprint arXiv:2206.01670}, 2022.

\bibitem{miech2019howto100m}
Antoine Miech, Dimitri Zhukov, Jean-Baptiste Alayrac, Makarand Tapaswi, Ivan
  Laptev, and Josef Sivic.
\newblock Howto100m: Learning a text-video embedding by watching hundred
  million narrated video clips.
\newblock In {\em ICCV}, pages 2630--2640, 2019.

\bibitem{distilbert}
Victor Sanh, Lysandre Debut, Julien Chaumond, and Thomas Wolf.
\newblock Distilbert, a distilled version of bert: smaller, faster, cheaper and
  lighter.
\newblock {\em arXiv preprint arXiv:1910.01108}, 2019.

\bibitem{wray2019fine}
Michael Wray, Diane Larlus, Gabriela Csurka, and Dima Damen.
\newblock Fine-grained action retrieval through multiple parts-of-speech
  embeddings.
\newblock In {\em ICCV}, pages 450--459, 2019.

\end{thebibliography}
